\documentclass[conference]{IEEEtran}
\IEEEoverridecommandlockouts
\usepackage{cite}
\usepackage{amsmath,amssymb,amsfonts}
\usepackage{algorithmic}
\usepackage{graphicx}
\usepackage{textcomp}
\def\BibTeX{{\rm B\kern-.05em{\sc i\kern-.025em b}\kern-.08em
    T\kern-.1667em\lower.7ex\hbox{E}\kern-.125emX}}

\usepackage{graphicx}

\usepackage{pgfplots}
\usepackage{tikz}
\usetikzlibrary{arrows,decorations.pathmorphing,fit,positioning}
\usepackage{xcolor}
\usetikzlibrary{patterns}
\usepackage{amstext}
\usepackage{pgf-pie}
\usepackage{float}
\usepackage{kbordermatrix}

\usepackage{multirow}

\usepackage[hyphenbreaks]{breakurl}
\usepackage[hyphens]{url}

\usepackage{tikz}
\usetikzlibrary{trees}

\usepackage{subfig}

\tikzset{level 1/.style={level distance=2cm, sibling distance=10cm}}
\tikzset{level 2/.style={level distance=2cm, sibling distance=3cm}}

\tikzset{bag/.style={text width=15em, text centered,yshift=-0.5cm}}

\usetikzlibrary{arrows,decorations.pathmorphing,fit,positioning}

\pgfplotsset{compat=1.5}

\def\addlegendimage{\csname pgfplots@addlegendimage\endcsname}

\begin{document}

\title{Taming Wild High Dimensional Text Data with a Fuzzy Lash
}

\author{\IEEEauthorblockN{Amir Karami}
	\IEEEauthorblockA{\textit{School of Library and Information Science} \\
		\textit{University of South Carolina}\\
		Columbia, SC, USA \\
		karami@sc.edu}
}

\maketitle

\begin{abstract}

The bag of words (BOW) represents a corpus in a matrix whose elements are the frequency of words. However, each row in the matrix is a very high-dimensional sparse vector. Dimension reduction (DR) is a popular method to address sparsity and high-dimensionality issues. Among different strategies to develop DR method, Unsupervised Feature Transformation (UFT) is a popular strategy to map all words on a new basis to represent BOW. The recent increase of text data and its challenges imply that DR area still needs new perspectives. Although a wide range of methods based on the UFT strategy has been developed, the fuzzy approach has not been considered for DR based on this strategy. This research investigates the application of fuzzy clustering as a DR method based on the UFT strategy to collapse BOW matrix to provide a lower-dimensional representation of documents instead of the words in a corpus. The quantitative evaluation shows that fuzzy clustering produces superior performance and features to \textit{Principal Components Analysis} (PCA) and \textit{Singular Value Decomposition} (SVD), two popular DR methods based on the UFT strategy.

\end{abstract}

\begin{IEEEkeywords}
dimension reduction, fuzzy clustering, SVD, PCA, classification
\end{IEEEkeywords}

\section{Introduction}

Large electronic archives provide extremely useful and valuable resources to the scholarly community \cite{karami2015fuzzyiconf}. For example, there are more than 25 million documents in the MEDLINE/PubMed website\footnote{\url{https://www.nlm.nih.gov/bsd/licensee/baselinestats.html}} and more than 4 million documents in the IEEE Xplore Digital Library website\footnote{\url{https://www.ieee.org/about/today/at_a_glance.html#sect1}}. This huge amount of documents has created a growing need to develop new methods for processing high dimensional data \cite{karami2015fuzzy}. This computational area is one of the data-intensive challenges identified by National Science Foundation (NSF) as an area for future study \cite{council2016future}.

Bag-of-words (BOW) is a common method in text data representation. This technique represents documents based on the frequency of words with a matrix \cite{karami2014fftm}. However, this high dimensional matrix is a sparse matrix for large number of documents \cite{karami2015flatm}. Sparsity means that most elements in BOW matrix are zero because each document contains a small percentage of all words in a corpus \cite{aggarwal2012introduction}.

Dimension reduction (DR) is a per-processing step for reducing the original BOW dimension. The objectives of dimension reduction strategies are to improve speed and accuracy of data mining \cite{karami2015fuzzy}. There are four main strategies for DR: Supervised-Feature Selection (SFS), Unsupervised-Feature Selection (UFS), Supervised-Feature Transformation (SFT), and Unsupervised-Feature Transformation (UFT) \cite{karami2015fuzzy}. Feature selection focuses on finding a feature subset that can describe the data, as good as the original dataset, for supervised or unsupervised learning tasks  \cite{wu2002feature}. Unsupervised means there is no teacher, in the form of class labels \cite{liu2007computational}. Many existing databases are unlabeled because large amounts of data make it difficult for humans to manually label the categories of each document. Moreover, human labeling is expensive and subjective. Hence, unsupervised learning is needed.

DR reduction methods are based on some approaches such as linear algebra, statistical distributions, and neural network. Fuzzy approach has contributed to decision making \cite{karami2010risk,karami2012fuzzy} and data mining in various ways by providing a flexible approach such as fuzzy information granulation and representing vague patterns \cite{hullermeier2011fuzzy}; however, fuzzy clustering has not been considered as a DR approach. 

This paper will discuss the application of fuzzy clustering for dimensionality reduction based on the UFT strategy. This research compares the DR performance of fuzzy clustering, PCA, and SVD, and shows that fuzzy clustering has better performance in document classification and has computational advantages over the current methods. 

The remainder of this paper is organized as follows. In the related work section, we review the DR research. In the methodology and experiment sections, we provide more details about using fuzzy clustering as a DR method along with an evaluation study to verify the effectiveness of fuzzy clustering. Finally, we present a summary, limitations, and future directions in the last section.

\section{Related Work}

Big text data have encouraged researchers to propose dimension reduction techniques in four categories \cite{cunningham2008dimension}: SFS, SFT, UFS, and UFT. 

SFS strategy explores the best minimum subset of the original words (features) for labeled data. Assume that $W=\{w_1,w_2,...,w_m\}$ and $L=\{l_1,l_2,...,l_p\}$ denote the words and the class label set where $m$ and $p$ are the number of words and labels, respectively. $D=\{d_1,d_2,...,d_n\}$ is the corpus where $n$ is the number of documents. The goal of SFS strategy is to find $F=\{f_1,f_2,...,f_k\}$ that is a subset of $W$ with $k$ features ($k < m$) with respect to $L$.  Several methods were developed based on SFS strategy such as information gain \cite{yang1997comparative} and Chi-square measure \cite{gao2017learning}.

SFT strategy maps the words to a new basis for labeled data. The goal of SFT strategy is to map the words in $W$ onto clusters, $C=\{c_1,c_2,...,c_k\}$, with respect to $L$ where $k << m$. For example, Linear Discriminant
Analysis (LDA) is a SFT method using Fisher criterion based on maximizing the between class scatter and minimizing the within class scatter \cite{mika1999fisher}. 

UFS explores the best minimum subset of the original words for unlabeled data. The goal of unsupervised-feature selection strategy is to find the best minimum subset ($k$) of $F$ without having $L$ where $k < m$.  Different methods have been developed based on UFS strategy such as Non-negative Matrix Factorization (NMF) \cite{lee1999learning} and Laplacian Score (LS) \cite{he2006laplacian}.

the UFT strategy maps the words to a new basis for unlabeled data. The goal of unsupervised-feature transformation strategy is to map the words in $W$ onto $C$ without having $L$ where $k << m$. Several methods have been developed based on the UFT strategy such as  Principal Components Analysis (PCA) that is a linear unsupervised-feature transformation to map a set of correlated features into a set of uncorrelated features using orthogonally \cite{abdi2010principal}. PCA is among the most effective dimension reduction techniques and has shown a better performance than other techniques \cite{van2009dimensionality}. While PCA uses eigen-decomposition of the covariance matrix, Latent Semantic Analysis (LSA) is a similar method using Singular Value Decomposition (SVD) for feature transformation \cite{deerwester1990indexing}. SVD detects the maximum variance of the data in a set of orthogonal basis vectors \cite{sweeney2014comparison}.

Some studies have applied fuzzy approach to develop dimensionality reduction methods based on supervised- and unsupervised- feature selection strategies such as Rough Set Attribute Reduction (RSAR) \cite{jensen2004semantics}. The current fuzzy-based dimension reduction methods rely on retaining important features, and removing irrelevant and redundant (noisy) features \cite{mac2013unsupervised}; however, this strategy loses some information. This research investigates the potential of fuzzy clustering as a DR method and compares its performance with powerful current DR methods based on the UFT strategy.

\section{Method}

The goal of UFT strategy is to obtain a new basis that is a combination of the original basis. Among different methods with respect to this strategy, PCA and LSA are well-known widely used methods \cite{hinton2006reducing}. PCA converts matrix $X$ that contains $n$ objects or documents with $m$ variables or words to three matrices: linear combination of variables for each object ($t$), vectors of regression coefficients ($P$), and residuals ($E$):

\begin{center}
$X= tP^T+E$
\end{center}

LSA applies SVD on matrix $X$ to drop  the least significant singular values and keep $k$ singular values. SVD converts matrix $X$ to three matrices: diagonalized $XX^T (U)$, singular values of $X (S)$, and diagonalized $X^TX (V^T)$. In both PCA and SVD, the original basis is represented by a new reduced base with k dimensions ($d<<m$ and $d<<n$):
 
\begin{center}
$X=USV^T$
\end{center}

The traditional reasoning has a precise character that is yes-or-no rather than more-or-less \cite{zimmermann2010fuzzy}. Fuzzy logic proposes a new approach to move from the classical logic, zero or one, to the truth values between zero and one \cite{zadeh1973outline,karami2012fuzzy}. 

Fuzzy logic assumes that if $X$ is a collection of data points represented by $x$, then a fuzzy set $A$ in $X$ is a set of order pairs, $ A=\{(x,\mu_A (x)|x \in X)\}$. $\mu_A(x)$ is the membership function which maps $X$ to the membership space $M$ which is between 0 and 1 \cite{karami2012fuzzy}.

The goal of most clustering algorithms is to minimize the objective function ($J$) that measures the quality of clusters to find the optimum $J$ which is the sum of the squared distances between each cluster center and each data point. There are two major clustering approaches: hard and fuzzy (soft) \cite{karami2017fuzzy}. The hard approach assigns exactly one cluster to a document, but the soft approach assign a degree of membership with respect to each of cluster for a document \cite{karami2015fuzzy}. Among fuzzy clustering techniques, fuzzy C-means (FCM) is the most popular model \cite{bezdek1981pattern} to minimize an objective function by considering constraints: 

\begin{equation}
Min \: \: J_q =\sum_{f=1}^{k} \sum_{j=1}^{n} (\mu_{fj})^q ||d_j-v_f||^2         
\end{equation}
subject to:
\begin{equation}
0 \leq \mu_{fj}\leq1;
\end{equation}
\begin{equation}
\sum_{f=1}^{c} \mu_{fj}=1
\end{equation}
\begin{equation}
0<\sum_{j=1}^{n} \mu_{fj} < n; 
\end{equation}
Where: \\

\noindent $n$= number of documents\\
$k$= number of clusters\\
$\mu$= membership value\\
$q$= fuzzifier, $1 < q \le \infty$ \\
$d$= document vector\\
$v$= cluster center vector\\

\begin{figure}[ht]
	\centering
	\begin{tikzpicture}
	\small
	
	\draw (-2.2,0) node [rotate=90] {Documents};
	\draw (0,1.2) node {Words};
	
	\draw (4,1.2) node {Fuzzy Clusters};
	
	\draw (2.75,0) node [rotate=90] {Documents};

	\draw (-2,-1) rectangle (2,1) node[pos=.5] {$X_{n \times m}$};
	\small \draw (2.4,0) node {$\rightarrow$};
	\tiny
	\draw (3,-1) rectangle (5,1) node[pos=.5] {$C_{n \times k}$};

	
	\end{tikzpicture}

	\caption{Matrix Interpretation of FC}
	\label{tab:FC}
\end{figure}
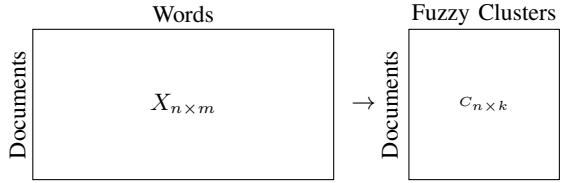

In this research, we use fuzzy clustering to find $\mu_{fj}$ as the membership degree for each document ($d_j$) with respect to each of clusters. The value of $\mu_{fj}$ is between 0 and 1 and is assumed to be a new basis to represent document-term frequency matrix. The number of documents and the number of clusters are represented by $n$ and $k$. We assume that fuzzy clustering converts $X$ with $n$ documents and $m$ words to a new reduced matrix (C) with $k$ variables or dimensions ($k<<m$) (Fig. \ref{tab:FC}). It is worth mentioning that fuzzy clustering does not lose information in $X$ and does not need to select a subset of dimensions such as SVD.

For example, assume that there are 10 words in a corpus with 5 documents represented by $X$ matrix whose elements show the frequency of the words in each of the documents. For instance, word 1 ($w_1$) is appeared two times in document 2 ($d_2$).   Applying fuzzy clustering on $X$ to find two fuzzy clusters creates matrix $C$ that each element is a cluster's membership degree with respect to a document.  For instance, document 1 ($d_1$) with $0.2118281$ membership value belongs to cluster 1 ($c_1$) and with  $0.7881719$ membership value belongs to cluster 2 (Fig. \ref{fig:fcexm}). In this example, fuzzy clustering converts $X_{5 \times 10}$ matrix to $C_{5 \times 2}$ matrix and reduces the dimension space by 80\%. 

\begin{figure*}[htp!]
	\[
	X= \kbordermatrix{
		& w_1 & w_2 & w_3 & w_4 & w_5 & w_6 & w_7 & w_8 & w_9 & w_{10} \\
		d_1 & 1& 0& 0& 1& 0& 0& 1& 2& 1&0\\
		d_2 & 2& 0& 1& 0& 0& 1& 0& 0& 0& 1 \\
		d_3 & 1& 0& 0& 2& 1& 0& 0& 1& 1&0 \\
		d_4 & 1& 1& 0& 0& 0& 1& 1& 1& 0&1 \\
		d_5 & 0& 0& 0& 1& 0& 1& 0& 0& 0& 0 
	} \rightarrow C=\kbordermatrix{
	& c_1 & c_2  \\
	d_1 & 0.2118281 & 0.7881719 \\ 
	d_2 & 0.8619096 & 0.1380904 \\
	d_3 & 0.0681949 & 0.9318051 \\
	d_4 & 0.8301873 & 0.1698127 \\
	d_5 & 0.4106981 & 0.5893019	
}
\]

\caption{A Numerical Example for Dimension Reduction Application of Fuzzy Clustering}
\label{fig:fcexm}
\end{figure*}

A large number of fuzzy clustering algorithms has been developed \cite{baraldi1999surveyI,baraldi1999surveyII}. To mange text data sparsity, we use a spherical fuzzy clustering, called soft spherical k-means. This method iterates between determining optimal memberships for fixed prototypes and computing optimal prototypes for fixed memberships \cite{dhillon2001concept}.

\section{Experiments}

In this section, we evaluate the dimension reduction application of fuzzy clustering against PCA and SVD by document classification using Functional Trees (FT), Random
Forest, and Adaptive Boosting (AdaBoost). that are among high performance classification algorithms \cite{chimieski2013association,sweeney2014comparison,rao2015performance,wu2008top,caruana2006empirical,qi2006evaluation}. 

We use two datasets, the irbla R package for computing SVD and PCA \cite{Baglama2017irlba}, the skmeans R package for soft (fuzzy) spherical k-means with 100 iterations and 1e-5 as the minimum improvement in objective function between two consecutive iterations \cite{Hornik2017skmeans}, and the Weka tool\footnote{\url{http://www.cs.waikato.ac.nz/ml/weka/}} with its default settings for the document classification.

\subsection{Datasets}

We leverage two datasets in this research:

\begin{itemize}
\item The Reuters dataset\footnote{https://archive.ics.uci.edu/ml/datasets/reuters-21578+text+categorization+collection} has 21,578 documents with several news categories. Two classes were created for binary classification. The documents in the Grain class were labeled as ``Grain" and the rest of the documents were labeled as ``Not Grain".

\item The Ohsumed dataset\footnote{http://disi.unitn.it/moschitti/corpora.htm} has 20,000 documents with different cardiovascular diseases categories. Two classes were created for binary classification. The documents in the Virus Diseases class were labeled as ``Virus Diseases" and  5000 documents randomly selected from the rest of the documents were labeled as ``Not Virus Diseases". 
\end{itemize}

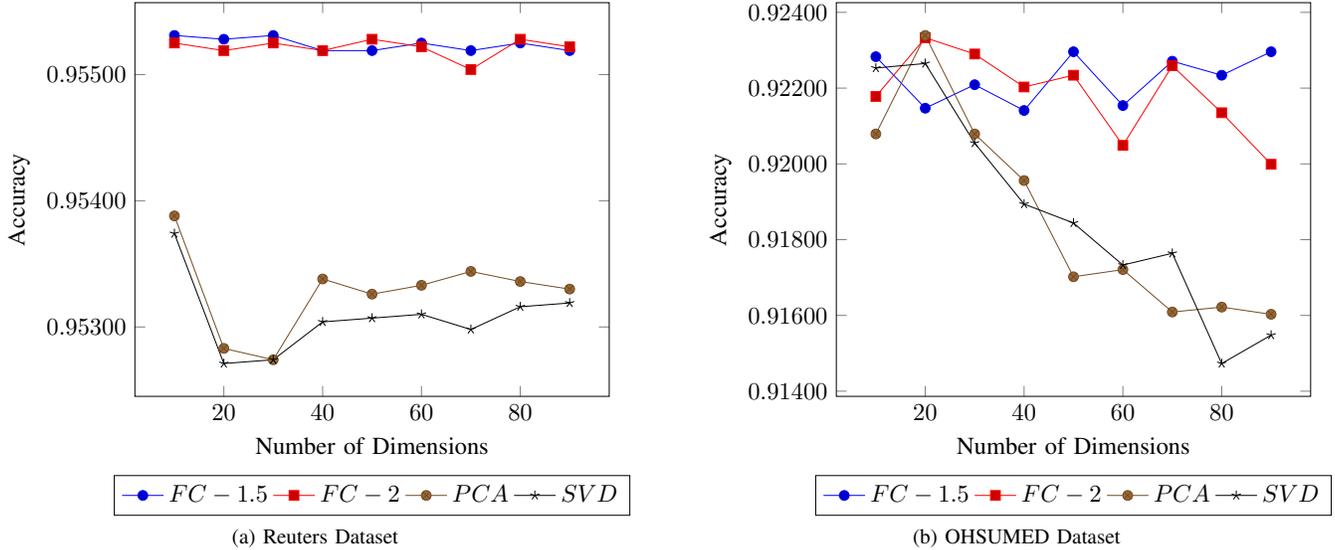
\begin{figure*}[htp!]
	\centering
	
	\subfloat[Reuters Dataset]{{
			\begin{tikzpicture}[scale=.92]
			\begin{axis}
			[xlabel=Number of Dimensions,ylabel= Accuracy,legend style={at={(0.5,-0.2)}, anchor=north,legend columns=-1},
			y tick label style={
				/pgf/number format/.cd,
				fixed,
				fixed zerofill,
				precision=5,
				/tikz/.cd
			}
			]
			
			\addplot coordinates { (10,0.95531) (20,0.95528) (30,0.95531) (40,0.95519)(50,0.95519)(60,0.95525)(70,0.95519)(80,0.95525)(90,0.95519)};

			\addplot coordinates { (10,0.95525)(20,0.95519)(30,0.95525) (40,0.95519)(50,0.95528)(60,0.95522) (70,0.95504) (80,0.95528)(90,0.95522)};

			\addplot coordinates { (10,0.95388) (20,0.95283) (30,0.95274)(40,0.95338)(50,0.95326)(60,0.95333)(70,0.95344)(80,0.95336) (90,0.95330)};
			
			\addplot coordinates { (10,0.95374)(20,0.95271)(30,0.95274)(40,0.95304)(50,0.95307)(60,0.95310) (70,0.95298) (80,0.95316) (90,0.95319)};
			
			\legend{$FC-1.5$,$FC-2$,$PCA$, $SVD$}
			
			\end{axis}
			\end{tikzpicture}				}}
	\qquad
	\subfloat[OHSUMED Dataset]{{
			\begin{tikzpicture}[scale=.92]
			\begin{axis}
			[xlabel=Number of Dimensions,ylabel= Accuracy,legend style={at={(0.5,-0.2)}, anchor=north,legend columns=-1},
			y tick label style={
				/pgf/number format/.cd,
				fixed,
				fixed zerofill,
				precision=5,
				/tikz/.cd
			}
			]
			\addplot coordinates { (10,0.92283)(20,0.92147)(30,0.92209)(40,0.92141) (50,0.92296) (60,0.92154)(70,0.92271)(80,0.92234) (90,0.92296)};
			
			\addplot coordinates { (10,0.92178)(20,0.92333)(30,0.92290) (40,0.92203) (50,0.92234) (60,0.92049)(70,0.92259)(80,0.92135) (90,0.91999)};

			\addplot coordinates { (10,0.92079) (20,0.92339)(30,0.92079)(40,0.91956) (50,0.91702) (60,0.91721)(70,0.91609)(80,0.91622) (90,0.91603)};
			
			\addplot coordinates { (10,0.92253)(20,0.92265)(30,0.92055)(40,0.91894) (50,0.91844) (60,0.91733)(70,0.91764)(80,0.91473) (90,0.91548)};
			
			\legend{$FC-1.5$,$FC-2$,$PCA$, $SVD$}
			\end{axis}
			
			\end{tikzpicture} }}

	\caption{Classification Evaluation}
	\label{fig:Acc}
\end{figure*}

\subsection{Document Classification}

Document classification problem assigns a document to a class. For this purpose, a pre-processing step is needed to extract features from text data. Using words in a corpus as features creates a large sparse matrix. One solution to reduce the feature set is to use DR methods such as fuzzy clustering, SVD, and PCA to reduce the number of the original features. 

Three classification methods including Functional Trees (FT), Random Forest, and Adaptive Boosting (AdaBoost) were trained on 10, 20, 30, 40, 50, 60, 70, 80, 90, 100 reduced dimensions. To avoid any possible sampling bias, we apply the 5-fold cross validation method that the data is broken into 5 subsets for 5 iterations. Each of the subsets is selected for testing and the rest of them are selected for training. 

The output of a classification method is presented as a confusion matrix (Table \ref{tab:confmx}) with the following definitions:

\begin{table}[H]
	\centering
	\caption{Confusion Matrix}
	\begin{tabular}{cc|c|c|c|}
		\cline{3-4}
		& & \multicolumn{2}{ c| }{\textbf{Predicted}} \\ \cline{3-4}
		& & \textbf{Negative} & \textbf{Positive}  \\ \cline{1-4}
		\multicolumn{1}{ |c| }{\multirow{2}{*}{\textbf{Actual}} } &
		\multicolumn{1}{ |c| }{\textbf{Negative}} & TN & FP    \\ \cline{2-4}
		\multicolumn{1}{ |c  }{}                        &
		\multicolumn{1}{ |c| }{\textbf{Positive}} & FN & TP      \\ \cline{1-4}
		
	\end{tabular}

	\label{tab:confmx}
\end{table}

\begin{itemize}
	\item True Negative (TN) is the number of correct predictions that an instance is negative.
	
	\item False Negative (FN) is the number of incorrect of predictions that an instance negative.
	
	\item False Positive (FP) is the number of incorrect predictions that an instance is positive.
	
	\item True Positive (TP) is the number of correct predictions that an instance is positive.
\end{itemize}

Classification accuracy of a classifier is an evaluation metric to measure how well the classifier recognizes instances of the various classes. The accuracy of a classifier is the percentage of correctly classified documents in a test set \cite{chimieski2013association}. 

\begin{equation} 
Accuracy = \frac{TP+TN}{TP+TN+FP+FN}
\end{equation}

\subsection{Evaluation Results}

Fig. \ref{fig:Acc}.a and Fig. \ref{fig:Acc}.b show the average of the accuracy of the three classifiers along with two fuzzifier values including 1.5 (FC-1.5) and 2 (FC-2) for the two datasets. These two figures indicate that fuzzy clustering illustrates better accuracy performance than PCA and SVD. 

In addition, FC-1.5 has better performance in most of the classification experiments and shows the highest stability with the lowest standard deviation value following by FC-2, SVD, and PCA. Although increasing the number of dimensions mostly has the negative effect on the accuracy performance of PCA and SVD based on Fig. \ref{fig:Acc}, fuzzy clustering shows a stable performance with lower standard deviation than the non-fuzzy ones. While SVD shows more stability than PCA, the latter one has better accuracy than the earlier one with different number of dimensions.

While the complexities for the PCA and the SVD methods are $O(mnlog(k))$ and $O(mnlog(k)+(m+n)k^2)$, respectively \cite{halko2009finding},  the complexity for the fuzzy spherical k-means is  $O(n+k)$ \cite{dhillon2002iterative}. Other than the complexity advantage, there are other benefits for the DR application of fuzzy clustering including not losing information,  estimating the number of clusters or dimensions with already developed methods such as silhouette index \cite{campello2006fuzzy} and Xie-Beni index \cite{xie1991validity}, and working with both discrete and continuous data.

\section{Conclusion}

The big text data databases represent extremely useful resources to the scholarly community; however, analyzing individual words in a corpus leads to a high dimensional sparse BOW matrix. DR is a pre-processing step in data mining to reduce BOW matrix dimension for better accuracy. Although a wide range of DR methods has been developed, the exponential growth of data indicates that DR still needs new perspectives. DR methods have been developed based on different strategies. UFT is a popular and efficient strategy using different approaches such as linear algebra, statistical distributions, and neural network. However,  fuzzy clustering has not been considered as a DR approach based on the UFT strategy. 

This study discusses the potential of fuzzy clustering for DR based on the UFT strategy. Fuzzy clustering processes BOW matrix and creates a new matrix whose elements are membership degree values for each document in a corpus. This research uses the new matrix as a reduced matrix of BOW matrix.  The efficiency and effectiveness of fuzzy clustering are demonstrated through accuracy comparisons with PCA and SVD using two public available corpora.

This paper's results illustrate that fuzzy clustering is a competitor to the powerful methods such as PCA and SVD in the setting of dimensionality reduction for document collections. Indeed, the principal advantages fuzzy clustering include not losing information and having less complexity.  Fuzzy clustering also works with both discrete and continuous data and there are developed methods to estimate the optimum number of dimensions. Although this paper has applied fuzzy clustering for text data dimension reduction purpose, this clustering method can be used for other data types such as image and microarray data.      

This research has several limitations. First, word weighting methods such as entropy are not considered. Second, the fuzzifier is limited to two values (1.5 and 2). Third, the accuracy improvement of the fuzzy clustering over PCA and SVD is not significant. In our future work, we will apply word weighting methods on fuzzy clustering, investigate different fuzzifier values, and explore other fuzzy clustering methods.

\bibliographystyle{IEEEtran}
\bibliography{refrence}

\end{document}